\def\year{2022}\relax
\documentclass[letterpaper]{article} 
\usepackage{aaai22}  
\usepackage{times}  
\usepackage{helvet}  
\usepackage{courier}  
\usepackage[hyphens]{url}  
\usepackage{graphicx} 
\usepackage{natbib}  
\usepackage{caption} 
\DeclareCaptionStyle{ruled}{labelfont=normalfont,labelsep=colon,strut=off} 
\usepackage{algorithm}
\usepackage{algorithmic}

\usepackage{array}
\usepackage{multirow}
\usepackage{caption}
\usepackage{subcaption}
\usepackage{color, colortbl}
\definecolor{Gray}{gray}{0.9}

\usepackage{newfloat}
\usepackage{listings}
\floatstyle{ruled}
\newfloat{listing}{tb}{lst}{}
\floatname{listing}{Listing}
\title{Privacy--Preserving Online Content Moderation: A Federated Learning Use Case}
\author{
    Pantelitsa Leonidou\textsuperscript{\rm 1}
    Nicolas Kourtellis \textsuperscript{\rm 2}
    Nikos Salamanos \textsuperscript{\rm 1}
    Michael Sirivianos \textsuperscript{\rm 1}
    }
\affiliations {
    \textsuperscript{\rm 1} Cyprus University of Technology  
    \textsuperscript{\rm 2} Telefonica Research \\
    pl.leonidou@edu.cut.ac.cy, nicolas.kourtellis@telefonica.com,  nik.salaman@cut.ac.cy, michael.sirivianos@cut.ac.cy
    }

\begin{document}

\maketitle

\begin{abstract}
Users are exposed to a large volume of harmful content that appears daily on various social network platforms. One solution to users' protection is developing online moderation tools using Machine Learning (ML) techniques for automatic detection or content filtering. On the other hand, the processing of user data by social network platforms requires compliance with privacy policies. Federated Learning (FL) is one of the proposed solutions, where the training of ML models is performed locally on the users’ devices (the FL ``clients''), and only the model updates are shared with the central server (the FL central aggregator). Although the raw data never leave the users' devices, privacy leaks can still occur. One threat is data (or membership) inference attacks, where an attacker accessing the final trained model (the FL output) can successfully perform unwanted inference of the data belonging to the users who participated in the training process. In this paper, we propose a privacy--preserving FL framework for online content moderation that incorporates Differential Privacy (DP). To demonstrate the feasibility of our approach, we focus on detecting harmful content on Twitter -- but the overall concept can be generalized to other types of misbehavior. We simulate a text classifier –- in a distributed FL fashion –- which can detect tweets with harmful content. We show that the performance of the proposed FL framework can be close to the centralized approach -- for both the DP and non--DP FL versions. Moreover, it has a high performance even if a small number of clients (each with a small number of data points) are available for the FL training. When reducing the number of clients (from fifty to ten) or the data points per client (from 1K to 0.1K), the classifier can still achieve ${\sim}{81}\%$ AUC. Furthermore, we extend the evaluation to four other Twitter datasets that capture different types of user misbehavior and still obtain a promising performance (61\% -- 80\% AUC). Finally, we explore the overhead on the users’ devices during the FL training phase and show that the local training does not introduce excessive CPU utilization and memory consumption overhead.

\end{abstract}

\section{Introduction}
Users of all ages are exposed to a large volume of information from various Online Social Networks (OSNs). The content is often questionable or even harmful regardless of age, expressing abusive behavior, extreme sarcasm, cyberbullying, racism, and offensive or hate speech. OSN platforms try to protect users by setting special terms and conditions, blocking malicious accounts, and flagging or even taking down harmful content. Despite these efforts, harmful content is still present. Researchers and developers have made a great effort to develop automated detection tools mainly based on Machine Learning (ML) algorithms ~\cite{founta2018large, 10.1145/2684822.2685316, Waseem2016HatefulSO, https://doi.org/10.48550/arxiv.1703.04009, chatzakou2017mean}. 
These ML models are first trained on large annotated datasets and then deployed online. One challenge is creating large labeled datasets suitable for deep learning training. The data are large (from millions of users), multi-modal (text, video, and images or a combination of those), and they change dynamically. Additionally, it is challenging for the platforms and researchers to collect and process these data in the first place. The users’ online data are private and sensitive, which is why the EU has imposed strict policies to protect users' privacy (GDPR and accompanying national legislation).

In this paper, we investigate whether privacy--preserving ML methods can effectively detect harmful online content while complying with privacy policies. For this purpose, we propose and evaluate a privacy--preserving Federated Learning (FL) framework for training text classifiers able to detect harmful content. FL is a collaborative ML training process where in each round, the training phase is performed locally at users' devices (the FL ``clients''), and only the model parameters are sent to the central server (the FL ``aggregator''). The central server aggregates the received information and updates the global model~\cite{mcmahan2017communicationefficient}. Therefore, FL has access to local, up-to-date user data and does not require such data to be globally collected by a central unit for storage and ML training; data that are usually massive in volume and a potential target for cyber-attacks, theft, and prying on.

Although the FL paradigm complies, in theory, with the GDPR policies (since the raw data never leave the users' devices), privacy leakages can still occur. Prior studies have shown that the FL framework is vulnerable to membership inference and backdoor attacks~\cite{NEURIPS2021_91cff01a, LDP_CDP}. In this paper, we consider Differential Privacy (DP) as a defense mechanism against membership and attribute inference attacks \cite{dwork2006, Dwork}. DP provides privacy guarantees (at the user level) against data (or membership) inference attacks by an external attacker who has access to the trained model. We incorporate the DP model proposed in~\cite{NEURIPS2021_91cff01a} which is a generalization of DP for the FL framework.

Our central research question is whether harmful online content can be detected efficiently and effectively by a privacy--preserving FL framework. To answer this, we bootstrap our ML text model from a modified version of the classifier presented in~\cite{founta2018unified}. We evaluate it when trained in an FL fashion (with and without DP) on different Twitter datasets from five studies of Twitter user misbehavior by generalizing the classification problem as detecting harmful or normal behavior.
We compare the classifier’s FL performance with the centralized version that has access to all data.
Finally, we assess a typical user device's overhead while training the classifier locally to examine whether the FL approach slows down the device.

This work makes the following contributions:
    \begin{itemize}
    \item We are the first to propose a methodology for applying privacy--preserving FL in the context of harmful content detection. Moreover, we provide a simulation methodology for using centralized datasets to test the performance of an FL framework. 
    
    \item We show that the performance of the proposed FL framework can be close to the centralized approach -- for both the DP and non--DP FL versions. The FL classification performance on a total of 100K tweets has only a 10\% difference in AUC compared to the centralized approach. For instance, by training the classifier (without DP) for only twenty FL rounds on fifty clients, we achieve ${\sim}{83}\%$ AUC. Moreover, when reducing the number of clients (from fifty to ten) or the data points per client (from 1K to 0.1K), the classifier can still achieve ${\sim}{81}\%$ AUC. In other words, we can achieve high performance even if few clients (with few data locally) are available.
        
    \item Our further evaluation of the classifier on four smaller Twitter datasets of other types of misbehavior shows promising performance, ranging from 61\% to 80\% AUC. This means that the classifier can generalize and detect different types of misbehavior.
    
    \item Finally, we show that the FL training process does not introduce excessive system overhead – in terms of CPU utilization and memory consumption –- on the users’ devices.
    
    \item The simulation and the classifier code are made available to the research community.
    \end{itemize}

\begin{figure*}[h!]
\centering
\includegraphics[width=0.90\textwidth]
{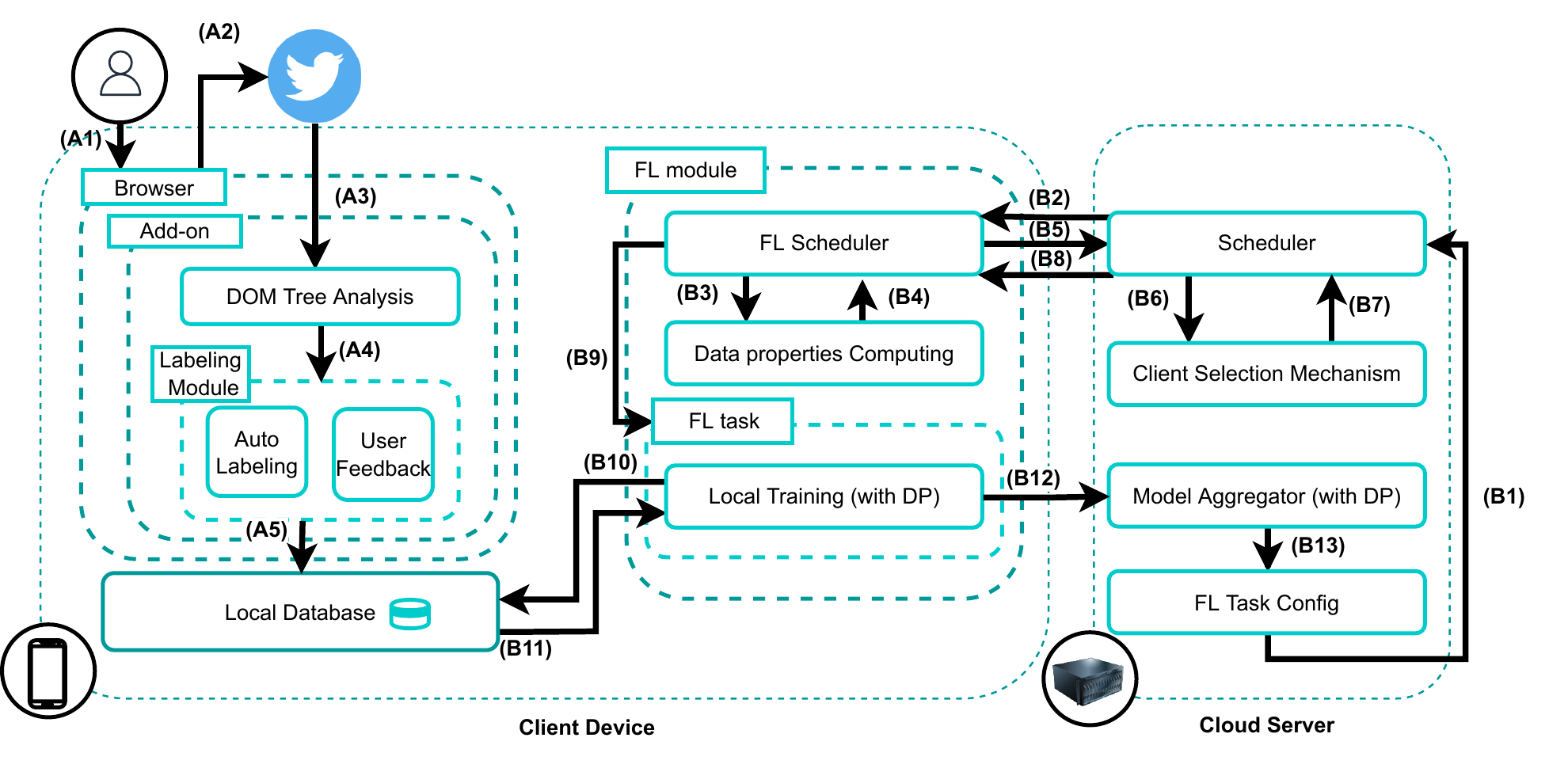} 
\caption{FL Differential Private Framework and Dataflow} 
\label{fig:framework}
\end{figure*}

\section{Related Work}

\subsection{Machine Learning for Automatic Detection and Filtering of Harmful Content}

Harmful content can be found in a text, visual (image, video), audio (songs, recordings) format, or a combination of those. We define any violent, abusive, sexual, disrespectful, hateful, illegal content, or any content that may harm the user as \textbf{``harmful''}. One solution to protect users from such content is adopting automatic detection or filtering using Machine Learning techniques in online moderation tools.

Several studies have investigated misbehavior on Twitter. \cite{chatzakou2017mean} proposes a deep-learning architecture to classify various types of abusive behavior (bullying and aggression) on Twitter. They proposed a methodology of extracting textual, user, and network-based features for Twitter accounts to identify patterns of abusive behavior. Then, they applied the methodology in a large dataset of 1.6M tweets collected during a period of three months. \cite{founta2018unified} presents a unified deep learning classifier to detect abusive texts on Twitter. The authors tested the unified classifier with several abusive Twitter datasets and achieved high performance. One of the evaluation datasets was the one presented in ~\cite{founta2018large} with 100K tweets labeled as ``Abusive'', ``Hate'', ``Normal'', and ``Spam'' using crowdsourcing annotation techniques. The unified classifier consists of two different classifiers whose results are combined to give the final result. One classifier is a text classification model, and the other treats domain-specific metadata (i.e., user's friend network, number of retweets, etc.). They tested the unified classifier with several abusive Twitter datasets and achieved high performance. In this work, we adopt a simplified version of the proposed classifier by replicating the model for the text classification task -- since we use no meta-data as training input but only text stored on a user's device. 

Yenala et al. proposed a deep learning architecture for detecting inappropriate language in query completion suggestions in search engines and users' conversations in messengers~\cite{yenala2018deep}. They used Convolution Neural Networks and Bi-directional LSTMs sequential model for the use case of the search engine suggestions and LSTM and Bidirectional LSTM sequential model for the users' conversations messengers. They prove that the suggested architecture outperforms pattern-based and hand-crafted feature-based architectures. The authors in \cite{10.1145/3442442.3452314} collected a dataset of ${\sim}{4}M$ records to assess the exposure of kids and adolescents to inappropriate comments on YouTube. They built a model consisting of five high-accuracy classifiers using Natural Language Processing and ML to classify the comments obtained in five age-inappropriate classes (Toxic, Obscene, Insult, Threat, Identity hate). The model acts as a binary classifier that classifies input as inappropriate if it falls into at least one of the five classes. 

Papadamou et al. built a deep learning classifier to detect videos with inappropriate content that targets toddlers on YouTube with high accuracy (84.3\%)~\cite{papadamou2019disturbed}. The authors in~\cite{9073060} created a dataset with three different categories of videos: ``Original Videos'', ``Explicit Fake Videos'', and ``Violent Fake Videos''. They trained a deep learning classifier to detect videos with content inappropriate for kids. with an accuracy of more than 90\%. Additionally, Papadamou et al. collected ${\sim}{7}K$ YouTube videos related to pseudoscientific content and used the resulting dataset to train a deep learning classifier to detect misinformation videos on YouTube and achieved an accuracy of 79\%~\cite{papadamou2021it}. These studies used video processing techniques to extract information from the videos but also collected other related information (i.e., the video's title, comments, caption, etc.). Moreover, this work focuses on detecting inappropriate text content.

\subsection{Federated Learning and Differential Privacy}
    McMahan et al. introduced Federated Learning(FL) as a distributed approach for training machine learning models without sharing an individual's data with a central unit \cite{mcmahan2017communicationefficient}. The idea is to train local models on clients' devices with their on-device available data and only share locally-computed updates with the central server. The server will collect the locally computed updates from the clients and aggregate them to update the global model. A client device in an FL setting can scale from a mobile device, a laptop, a desktop, or an IoT device to a company's data server.

    Since the FL appearance, many studies have described FL applications in real settings. Gboard \cite{yang2018applied} uses FL for training, evaluating, and deploying a model for giving optimized web, GIFs, and Stickers query suggestions. Gboard also used FL to train a model for next-word prediction\cite{hard2018federated}. Next word prediction is used on the keyboard to suggest words for the user to type next based on the text already typed. In \cite{DBLP:journals/corr/abs-1903-10635}, the authors applied FL to train a neural network to learn out-of-vocabulary (OOV) words to minimize annoying users by auto-correcting the OOV words considering them as misspellings. FL is also used to train an image-classification model to decide whether a patient has the COVID-19 virus or not using x-ray images from several hospitals to preserve the patients' privacy in~\cite{liu2020experiments}. The performance obtained when training the models using FL was slightly worse than training using a centralized approach.

    Several studies have shown that maintaining the raw data locally does not sufficiently protect the users' privacy so data leakages can occur in the FL framework. There are two main potential threats to data privacy; data inference attacks performed (i) by the other clients -- or even the central aggregator -- during the training phase and (ii) by an external attacker who has access to the final trained model. One of the proposed ways to provide privacy-preserving guarantees to FL is Differential Privacy (DP). The DP was first introduced by~\cite{dwork2006, Dwork} as a privacy--preserving technique for learning tasks on statistical databases. It can limit the information leakages regarding the data records used for the learning phase. DP provides statistical guarantees against data inference attacks performed by an adversary who has access to the output of the learning algorithm. These privacy guarantees are achieved by adding noise to the learning process to limit the data records' influence on the algorithm's final output. Two main variations of DP methodology have been incorporated into the FL framework toward privacy--preserving FL: the Central Differential Privacy (CDP) and the Local Differential Privacy (LDP)~\cite{LDP_CDP}; other hybrid approaches have also lately proposed~\cite{Hierarchical_Federated_Learning}. In CDP, the agents send the model updates to the central server, which will perform the DP noise addition~\cite{NEURIPS2021_91cff01a}. This implies that the central server is a trusted system entity, namely, it will not perform malicious inferences on the clients' data. In LDP, the DP noise addition is performed locally by the clients -- before sending the updates to the central server~\cite{LDP-Fed}. In this context, no trusted entity is required.

\section{Conceptual Framework}

To further explain the idea of applying the FL paradigm to the online moderation tools, we present our conceptual framework in Figure~\ref{fig:framework}. Regarding the \textbf{threat model} we assume that the only trusted entity is the central aggregator. Under the Central Differential Private protocol~\cite{NEURIPS2021_91cff01a} -- that we use in this study -- the central aggregator is responsible for adding the DP--noise on the model updates that receives from the clients in an FL round. This implies that the aggregator is a trusted entity, but the other participants may not. Hence, possible adversaries are either some clients or an external entity that may perform data inference attacks either during the training phase or through the final model. 

\subsection{System Components}

\textbf{Client Device:} The user's device that accesses the Online Social Network application (i.e, Twitter).

\noindent\textbf{Browser Add-On:} filters the users' online activity, conducts DOM tree analysis, and sends the selected data (i.e., tweets) to the Labeling Module.

\noindent\textbf{Labeling Module:} aggregates the labels obtained from the Auto-Labeling and User Feedback modules. \textit{Auto-Labeling module} can use semi-supervised learning techniques to label the data automatically. The \textit{User Feedback module} asks the user to label the data.

\noindent\textbf{Local Database:} stores the labeled data.

\noindent\textbf{FL Module:} schedules and executes FL tasks on the user device. \textbf{FL task} defines and executes the \textbf{Local training}.

\noindent\textbf{Data properties computing:} the module that computes the metadata of the user's dataset (i.e., size of data, etc.), accompanied by other device information (e.g., battery, internet connection type, device capabilities, etc.).

\noindent\textbf{Cloud Server:} a unit owned by a trusted party that coordinates the FL training.

\noindent\textbf{FL Task Configuration:} generates the FL Task description, which contains the baseline model for training, the criteria for the clients to participate in this task, and the FL parameters (e.g., number of FL rounds, the number of clients to participate, etc.).

\noindent\textbf{Scheduler:} advertise the FL task to the available clients and manage the communication with the clients.

\noindent\textbf{Client Selection Mechanism:} checks if the client's device complies with the criteria set by the FL Task Config module.

\noindent\textbf{Model Aggregator:} aggregates the clients' model updates and applies the aggregated update to the global model.

\subsection{Data-Flow}

Figure~\ref{fig:framework} shows the data flow of the proposed framework. Specifically:

\textbf{(A1)} The user accesses Twitter through the device's browser, \textbf{(A2)} and sends an HTTP request to Twitter. \textbf{(A3)} The \textit{Browser Add--On}'s \textit{DOM Tree Analysis module} receives the Twitter newsfeed page DOM tree, filters the user activity, and selects data for labeling. \textbf{(A4)} The \textit{Labeling module} receives the data (e.g. a tweet text). The \textit{Auto Labeling module} automatically labels the data. The \textit{User Feedback module} asks the user to label it. \textbf{(A5)} Then, the two labeling modules send the \{tweet, label\} pairs to the \textit{Labeling Aggregator}, which defines a final label for the tweet using an aggregation method, and \textbf{(A6)} stores the labeled data at the \textit{Local Database}.

When there is a pending FL task at the server, \textbf{(B1)} the \textit{FL Task Config module} sends the task description to the \textit{Scheduler}. \textbf{(B2)} The \textit{Scheduler} sends the task descriptions to the available clients. \textbf{(B3)} The \textit{client's FL Scheduler} receives it, and forwards it to the \textit{Data properties Computing module}, \textbf{(B4)} which sends the device properties back to it. \textbf{(B5)} The \textit{FL scheduler} sends the properties to the \textit{Scheduler}, \textbf{(B6)} which forwards them to the \textit{Client Selection Mechanism} to tell if the client will participate in the training or not. \textbf{(B7)} The mechanism module sends its positive or negative decision to the \textit{Scheduler}, \textbf{(B8)} which announces to the \textit{client's FL scheduler} its participation in training with the global model to train or closes the connection with it.

For participating clients, \textbf{(B9)} the \textit{FL Scheduler} sends the global model, and the task description to the \textit{FL Task module}, and \textbf{(B10)} requests the local dataset. \textbf{(B11)} The \textit{Local Database} sends the dataset, and starts the local ML training with Differential Privacy (DP) Adaptive Clipping. The \textit{Adaptive Clipping} receives the local model's updates, clip them, and \textbf{(B12)} sends them to the \textit{Model Aggregator}. The \textit{Model Aggregator} aggregates the updates, adds \textit{DP--noise} (i.e., it adds noise to the updates' sum), and applies the updates to the global model. Finally, \textbf{(B13)} it sends the model to the \textit{FL Task Config module} for its use in the next round of the FL training.

\begin{figure}
\centering
\includegraphics[width=\columnwidth]
{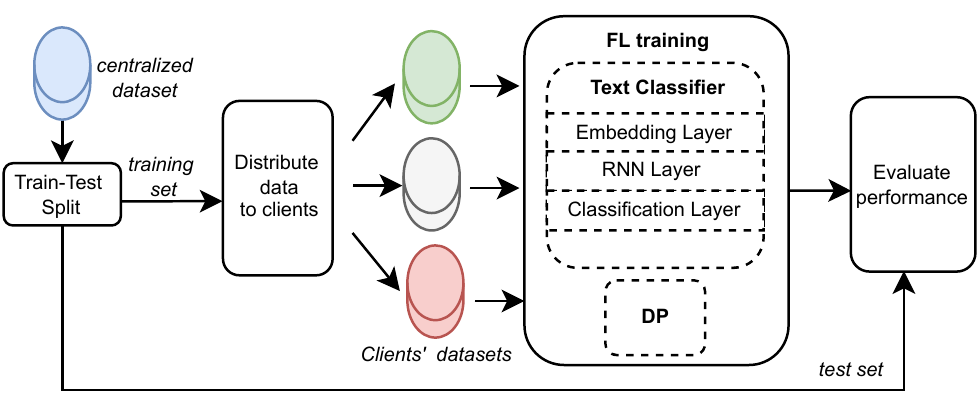} 
\caption{Federated Learning Setup Pipeline} 
\label{fig:pipeline}
\end{figure}

\section{Federated Learning Setup}

\subsection{General Assumptions}
Since we do not have access to the raw Twitter data from millions of users, the true distribution of harmful tweets to users is unknown. 
Thus, we have to somehow simulate the users' browsing history. For this purpose, we construct artificial clients by splitting a centralized Twitter dataset -- which contains harmful tweets -- into a number of disjoint sets. Moreover, we assume a homogeneous population of clients, namely, all clients have the same number of total tweets with the same ratio harmful to normal (i.e., that same class ratio). Additionally, we assume that the clients selected for the FL training remain available during the whole FL phase.

\subsection{Federated Learning Training}
We used TensorFlow Federated (TFF), an open--source framework for computations on decentralized data\footnote{\url{https://www.tensorflow.org/federated}}, to simulate the FL training process for our experiments. The FL algorithm we used for aggregating the client's model updates is the Federated Averaging~\cite{mcmahan2017communicationefficient}. TFF provides the implementation\footnote{\url{https://github.com/tensorflow/privacy}} of Differential Privacy for FL training which we use to add CDP in our FL training simulation. Figure~\ref{fig:pipeline} presents our pipeline to simulate the FL training. We describe next the FL pipeline's steps and main components. 

\subsection{Text classifier}
\label{sec:text-classifier-model}
We use a simplified version of the unified classification model described in~\cite{founta2018unified}, where only the text-classification path is enabled. We used this classifier since it showed a high performance ($\sim$80\% to $\sim$93\% AUC) across many harmful tweet datasets. We used this simplified version to give a lighter computational task to the user's device. The input of the classifier is the text of the tweets. We used TensorFlow Keras for the implementation of the classifier. The sequential ML pipeline starts with an Embedding layer, we use the GloVe embedding~\cite{pennington2014glove} with the highest dimension (200). A Recurrent Neural Network Layer follows with gated recurrent unit (GRU), 128 units, and a dropout of p=0.5. The output layer is a classification dense layer, with one neuron with the sigmoid activation function. We set the parameters as proposed in~\cite{founta2018unified}. TFF framework offers a function that wraps a Keras model\footnote{\url{https://www.tensorflow.org/federated/api_docs/python/tff/learning/from_keras_model}} for its use in the federated training simulation.

\subsection{Creating artificial clients for FL}
\label{sec:distributing_dataset}
We needed a decentralized dataset with a sufficient number of harmful and normal texts to simulate the FL training of the text classifier. Since we could not find a dataset fulfilling our criteria, we converted existing centralized datasets from past studies into artificial federated datasets. For this purpose, given a dataset with two classes of tweets (harmful and normal) and a sufficient number of harmful tweets, we do the following:

First, we create a test set with a size the 10\% of the dataset, with the condition that 8\% of the tweets in the test set are harmful. In other words, the class ratio harmful:normal in the test set is 8:92. We apply this percentage (8\%) based on the results of previous studies~\cite{founta2018large,chatzakou2017mean} that showed that the percentage of harmful content on Twitter is around ${\sim}{8}\%$.
Then, we create the clients using the remaining 90\% of the dataset. In our simulation, the clients are represented by sets of tweets (the clients' local data). To evaluate the FL on different populations of clients, we control the class ratio in clients' data, i.e harmful:normal. We also set the total number of tweets per client. Finally, given the clients' class ratio and clients' data size, we compute the maximum number of clients we can construct.

\section{Experimental Evaluation}

\subsection{Training Setup}\label{sec:training_setup}
To address the research questions of this work, we conducted experiments having the following training setups:\\
\textbf{FL training:} For the FL training setup, we are following the method described in Section~\ref{sec:distributing_dataset} -- given the parameters (clients' data size, percentage of harmful tweets) -- to construct the federated dataset. 
Then, we set the FL rounds and the number of participating clients in each round. Finally, we use the TFF framework to simulate the FL training. We refer to \textit{Local training} as the training of the model on the client's device, using the whole client's dataset as the local training set.

\noindent\textbf{Centralized training:} This is the traditional ML training setup where the text classifier is trained with a single train set. Regarding the train--test split, we construct the test set following the same procedure described in Section~\ref{sec:distributing_dataset}. That is, we initially split the dataset into a test set of 10\% size with class ratio 8:92 (i.e. 8\% harmful tweets). Then, from the remaining 90\% of the dataset, we construct the train set. We set a class ratio and a training--set size and then we randomly select a subset of tweets that satisfies these properties. 

In both setups, we train the text classifier described in Section~\ref{sec:text-classifier-model}, and we compute the weighted classification metrics\footnote{\url{https://scikit-learn.org/}}. We set the parameters (epochs=7, batch size=10, Adams optimizer, learning rate=0.001) after experimenting with different values for tuning and applying early stopping.
We run all the experiments on a server with Intel(R) Core(TM) i7-7700K CPU $@$ 4.20GHz, and a 62GiB RAM except for the ``overhead on client's device'' (Section~\ref{sec:overhead}) which we run on a Dell laptop device with Intel(R) Core(TM) i7-6500U CPU $@$ 2.50 GHz and 8GB RAM.

\subsection{FL simulation parameters}
The FL evaluation is based on the following three simulation parameters:

\begin{itemize}
    \item \textbf{Size of harmful class in each client:} With this parameter, we control the size of the harmful class on each client's dataset. We consider a homogeneous population with the same class ratio (harmful:normal). Generally, as studies showed, ${\sim}{8}\%$ of Twitter's online content is harmful~\cite{chatzakou2017mean,founta2018large}. That said, there are often controversial topics where the users' behavior is highly polarized. For instance, COVID-19 vaccination, the Russian invasion of Ukraine, and several conspiracy theories. We expect that the browsing history of users interested in these topics will contain a higher number of harmful content. 
        
    \item \textbf{Client dataset size:} the number of tweets at a client device. These tweets can represent either the user's browsing history or tweets posted, retweeted, etc., by the user.
    
    \item \textbf{Number of FL clients:} the number of clients available for the FL training. 
    \end{itemize}

We experiment with different values of the simulation parameters to explore how they affect the FL classification performance.

\subsection{Datasets}\label{sec:datasets}

\begin{figure*}{}
     \centering
     \begin{subfigure}[b]{0.32\textwidth}
         \centering
         \includegraphics[width=\textwidth, height=4cm]{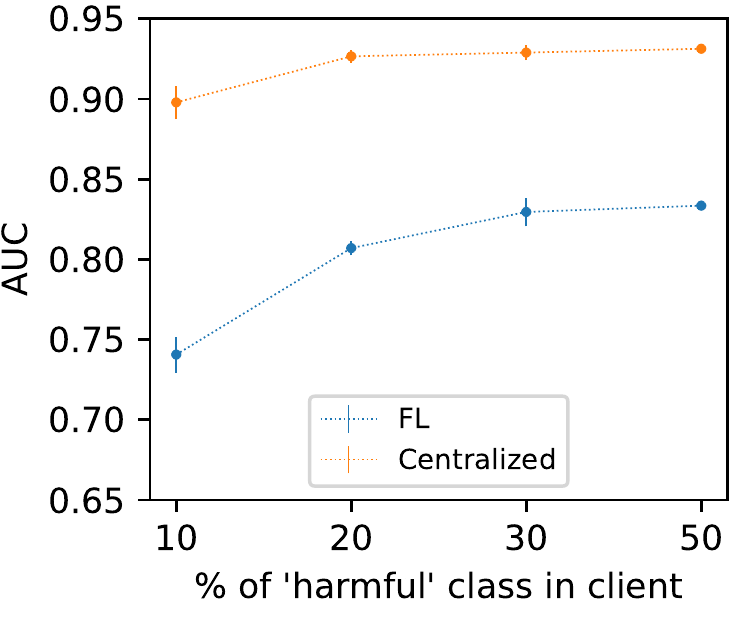}
         \caption{Harmful Class Ratio}
         \label{fig:class-ratio}
     \end{subfigure}
     \hfill
     \begin{subfigure}[b]{0.32\textwidth}
         \centering
         \includegraphics[width=\textwidth,height=4cm]{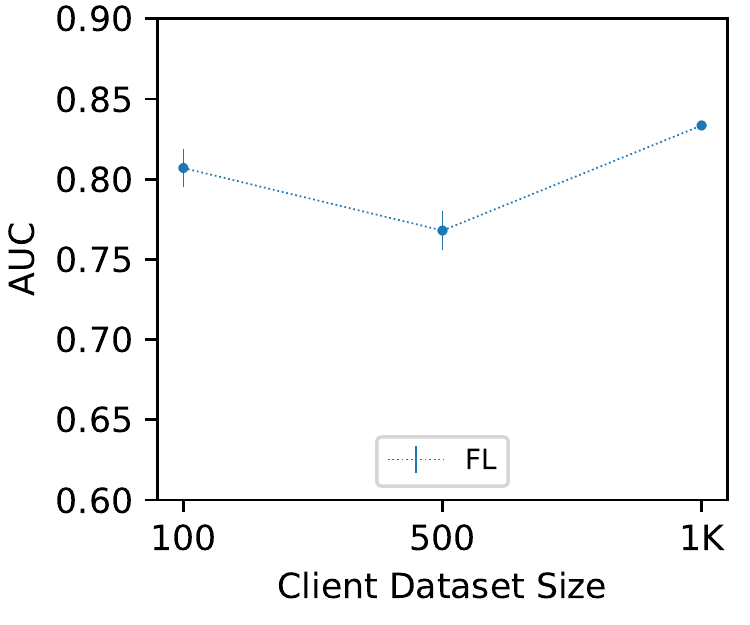}
         \caption{Client Dataset Size}
         \label{fig:client-size}
     \end{subfigure}
     \hfill
     \begin{subfigure}[b]{0.32\textwidth}
         \centering
         \includegraphics[width=\textwidth,height=4cm]{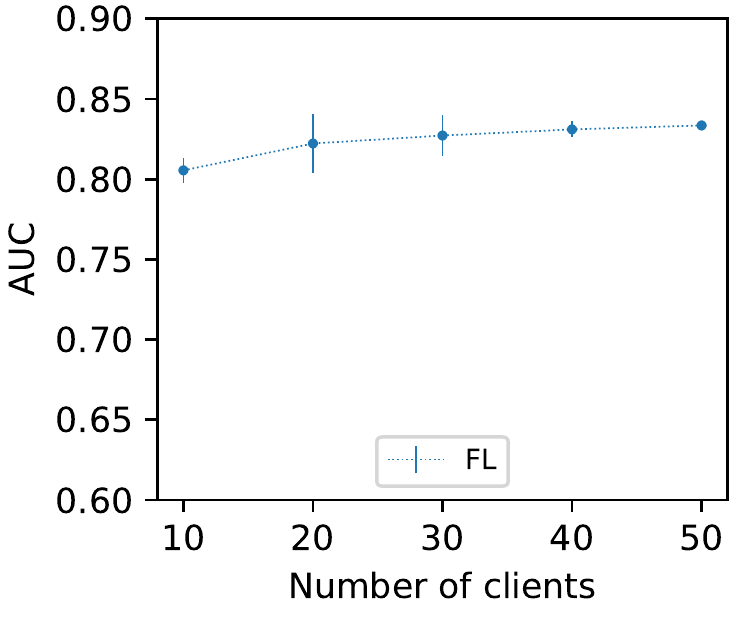}
         \caption{Number of Clients}
         \label{fig:number-clients}
     \end{subfigure}
        \caption{Evaluation of non-DP FL. (a) 50 clients, 1K data points per client; (b) 50 clients, balanced data per client (i.e. 50\% harmful data); (c) 1K data points and balanced data per client}
        \label{fig:auc-plots}
\end{figure*}

We select the following datasets for the experimental evaluation based on past studies of misbehavior on Twitter. For all datasets, in order to keep the FL task lighter for the user device, we binarize the classification problem by merging the several harmful classes into a single ``harmful'' class. We report below the original classes together with the final binary ones.

\noindent\textbf{Abusive Dataset}~\cite{founta2018large} initially contains ${\sim}{100}K$ tweets, labeled as ``Abusive'', ``Hate'', ``Normal'', and ``Spam''. We remove 14,030 tweets labeled as ``Spam'' -- following the same methodology of~\cite{founta2018unified} because there are more sophisticated techniques to handle spam profiles.
The resulting dataset consists of ${\sim}{86}K$ tweets with 31.6\% ``Abusive'', 5.8\% ``Hate'', and 62.6\% ``Normal'' classes. Final binary classes: 37.4\% ``Harmful'' and 62.6\% ``Normal''.

\noindent\textbf{Sarcastic Dataset}~\cite{10.1145/2684822.2685316} contains ${\sim}{61}K$ tweets text classified in two classes labeled as ``Sarcastic''(10.5\%), and ``None''(89.5\%). Final binary classes: 10.5\% ``Harmful'' and 89.5\% ``Normal''

\noindent\textbf{Hateful Dataset}~\cite{Waseem2016HatefulSO} is a ${\sim}{16}K$ tweets dataset. The tweets are categorized in ``Racism''(12\%), ``Sexism''(20\%), and ``Normal''(68\%) classes. Final binary classes: 32\% ``Harmful'' and 68\% ``Normal''.

\noindent\textbf{Offensive Dataset}~\cite{https://doi.org/10.48550/arxiv.1703.04009} consists of ${\sim}{25}K$ tweets categorized in three classes: ``Hate''(6\%), ``Offensive''(77\%), and ``Normal''(17\%).
Final binary classes: 83\% ``Harmful'' and 17\% ``Normal''.

\noindent\textbf{Cyberbully Dataset}~\cite{chatzakou2017mean} is a smaller dataset, with ${\sim}{6}K$ tweets distinguished the ``Bully''(8.5\%), ``Aggressive''(5.5\%), and ``Normal''(86\%) classes. Final binary classes: 14\% ``Harmful'' and 86\% ``Normal''.

We also \textbf{preprocess the tweet texts} by removing tags, URLs, numbers, punctuation characters, non-ASCII characters, etc.
Moreover, we convert the text to lowercase, all the white spaces into a single one. We also remove English stop words and words that appear only once in the dataset (in case of misspelled words).

\subsection{Evaluation of non--DP FL models} \label{sec:non-dp-experiments} 
In the following experiments, we evaluate the non--DP FL framework on the ``Abusive'' dataset only. We chose this dataset because its size allowed experimentation with various FL simulation parameters. 

\noindent\textbf{5.4.1 Percent of ``harmful'' data in the clients.}\label{sec:ratio}
Here, we evaluate the FL classification when we vary the percent of ``harmful'' data in the clients' datasets using the values $(10\%, 20\%, 30\%, 50\%)$. For a given ``\%harmful'' value, first, we randomly select fifty clients and then we train the classifier in these clients for twenty FL rounds. Each client dataset consists of 1K data. Finally, we repeat the experiment five times to acquire average scores. 

We also ran experiments with the \textit{Centralized training setup} by varying the percent of ``harmful'' text in the training set. Then, we randomly select 50K tweets as the training set.
We chose the 50K samples to compare the centralized classification performance with the previously mentioned FL training. We repeated the training three times for each \%harmful value.

\textbf{\textit{Results Discussion:}}
In Figure~\ref{fig:class-ratio}, we present the average AUC values (test evaluation).
We note that by increasing the examples of the ``harmful'' class by 5 times (i.e., from 10\% to 50\%), we have a ${\sim}$9\% increase in AUC (from 74\% to 83\%). In the case with a 10\% harmful class size, we got a 95\% score in precision, recall, and F1-score. Interestingly, in the case of 50\% of harmful class size, we obtained precision (93\%), recall (89\%), F1--score (90\%), which shows a decrease by ${\sim}$1\%, 6\%, and 4\% respectively. The training dataset is imbalanced when only 10\% of clients' data is harmful.
To understand this reduction in the model's performance, we also calculated the metrics only on the harmful class (which is the minority class), where we observed a ${\sim}$30\% increase in recall but also a 40\% negative impact on precision (with 10\% of harmful class size we got a recall of 50\%, and precision of 82\%, with 50\% we got a 77\%, and a 40\% respectively). 
This means having a balanced dataset (with 50\% of harmful class size) impacts the recall of the harmful class: i.e., it helps the model to learn better the harmful class. This is what drives AUC up as well (in the weighted metrics as well as in the harmful--only case). 

In the centralized approach, the classifier shows high performance, with only a 3\% AUC difference between the 10\% and 50\% of harmful class size (90\%, and 93\% AUC, respectively).
Finally, we get the best FL classification performance for balanced clients datasets (only ${\sim}$10\% AUC difference with the centralized training).

\noindent\textbf{5.4.2 Client's dataset size.} \label{sec:size}
We assumed a homogeneous setting where all clients have the same dataset size. We evaluate the classifier performance for the client's dataset size in [0.1K, 0.5K, 1K]. We run the \textit{FL training setup} for twenty FL rounds by using the same randomly selected fifty clients. Each client has a balanced dataset. We repeat the FL training twenty times for the training with 0.1K and 0.5K data, and five times for the 1K data. We present the average AUC metric in Figure~\ref{fig:client-size}.

\textbf{\textit{Results discussion:}}
Increasing client dataset size by ten times (from 0.1K to 1K data points) can lead to the overall improvement of performance metrics by ${\sim}{3}$\% in the AUC (from ${\sim}81$\% to 83\%). We observed also a ${\sim}{2}$\% improvement in F1 score (from 88\% to 90\%), ${\sim}{4}\%$ in accuracy (from 85\% to 89\%), recall (from 85\% to 89\%), and ${\sim}{1}\%$ in precision (from 92\% to 93\%).
The results show that increasing the data by five times did not significantly improve the performance, but the model performs similarly with the 0.1K data points per client.
Therefore, the experiment shows that the FL training can build an effective model (${\sim}{81}\%$ AUC) even with 100 data points per client.

\noindent \textbf{5.4.3 Number of FL Clients.} \label{sec:number_of_clients}
In this experiment, we run the \textit{FL training setup} by varying the number of available clients, i.e., 10, 20, 30, 40, 50.
Each client has a 1K balanced dataset, and the FL training runs for twenty rounds with the same randomly selected clients. We run the FL training five times for each value of the number of clients property, and we present the average test AUC in Figure~\ref{fig:number-clients}.

\textit{\textbf{Results Discussion:}} Increasing the number of clients participating in FL training by five times (i.e., from 10 to 50) results in increasing the AUC by ${\sim}2\%$ (from 81\% to 83\%). Additionally, the accuracy, precision, recall, and f1-score, increase by ${\sim}$3\%, 1\%, 3\%, and 2\% respectively (from 86\%, 92\%,86\%, 88\% to 89\%, 93\%, 89\%, 90\%).
However, the interesting point is that even with ten users/clients, the system can build an efficient model.
The model performs similarly well when varying the number of clients participating in the FL training.

\begin{table}
\centering
    \begin{tabular}{p{1.4cm}|p{1.1cm} c c c c c }
        \hline
        Dataset &  \#Clients  &  Accuracy &    AUC &  F1 Score\\
        \hline
        \multirow{4}{5em}{Abusive}  &  \multirow{2}{3em}{50}&   0.85 &        0.81 &      0.88  \\
       
        &      &              (0.01) &      (0.01) &      (0.01)  \\\cline{2-5}
        &  \multirow{2}{5em}{Centr.}	&0.92		&0.92	&0.94	\\
        
        &	 & (0.01)		&($<$1e-3)	&(0.01)	\\
        \hline
        \multirow{4}{3em}{Sarcastic} & \multirow{2}{3em}{50}    &      0.73 &     0.66 &  0.79\\ 
        &   &      (0.01) &        (0.01) &      (0.01)  \\\cline{2-5}
        
        &  \multirow{2}{5em}{Centr.} &	0.76 &	 0.75	&.0.83	\\
        & & (0.05)		& (0.03)	& (0.03)	\\
        \hline
        \multirow{4}{5em}{Hateful}&  \multirow{2}{3em}{50}   &      0.85 &        0.61 &      0.87 \\
        &                 &      (0.02) &        (0.01) &      (0.01) \\ \cline{2-5}
        &  \multirow{2}{5em}{Centr.}	& 0.79		& 0.79	&0.85 \\
        &	& (0.02) &	 (0.01)	& (0.01) \\
        \hline 
        \multirow{4}{5em}{Offensive}& \multirow{2}{3em}{37}  &      0.78 &      0.78 &      0.83  \\
        &  &         (0.02) &       (0.01) &      (0.02)  \\\cline{2-5}
        &  \multirow{2}{3em}{Centr.}	&0.92		&0.92	&0.94	\\
        & &(0.01)		&($<$1e-3)	&(0.01)	\\
        \hline
        \multirow{4}{5em}{Cyberbully}   &  \multirow{2}{3em}{16}   &      0.94 &        0.80 &      0.94  \\
        &    &      ($<$1e-3) &       (0.01) &      ($<$1e-3)  \\\cline{2-5}
        &   \multirow{2}{3em}{Centr.}	&0.91	&0.91	&0.93	\\
    	& &(0.03)		&(0.02)	&(0.02)	\\
     \hline
    \end{tabular}   
    \caption{Comparing FL and centralized approach. Average values over five repetitions (std in parenthesis) for five different datasets. Each client has 0.1K data points and balanced data (50\% harmful class).}
    \label{table:data}
\end{table}

\subsection{Generalization on other Twitter datasets}

Bootstrapping from the first round of experiments, we test the \textit{FL training setup} with four other datasets (see datasets details in Section~\ref{sec:datasets}) to explore the generalization of the classifier's utility. For each dataset, we run both the FL, and centralized training for five repetitions each, and then compare the average performances.

We run the FL training for twenty rounds, with the same clients participating in each round. Each client had a 100 tweets balanced dataset. We set the data size to 100 due to the datasets' size limitations, and based on the previous experiments that 100 data points per client are sufficient for effective FL training.
We randomly select fifty clients when the dataset size allowed us to do so. For small datasets we build the maximum number of clients i.e., 37 and 16 clients for Offensive and Cyberbully datasets, respectively. For the Centralized training, we used a training set size$=\#clients \times 100$ to fit the total data used in the FL training for the corresponding dataset. We did not perform hyperparameter tuning to train the model with the different datasets. We present the average evaluation metrics (test phase) for both setups in Table~\ref{table:data}.

\textit{\textbf{Results Discussion:}}
Across all five datasets, we observe an AUC performance ${>}{61}\%$. We get the best AUC while training with the Abusive dataset (81\%), and with the smallest dataset, the Cyberbully, we achieved an AUC of 80\%. Training with the Offensive, Sarcastic, and Hateful, we got an AUC performance of 78\%, 66\%, and 61\%, respectively. 
Additionally, we can observe that the model's performance decreases by ${\sim}{9}\%$ (the minimum) to ${\sim}{18}\%$ (the maximum) when trained with the FL approach compared to the centralized one. However, the results show that the classifier can be generalized and achieve acceptable performance on different types of misbehavior, even without hyperparameter tuning.

\subsection{FL with Central Differential Privacy}

\begin{figure*}[h]
     \centering
     \begin{subfigure}[b]{0.35\textwidth}
         \centering
         \includegraphics[width=\textwidth]{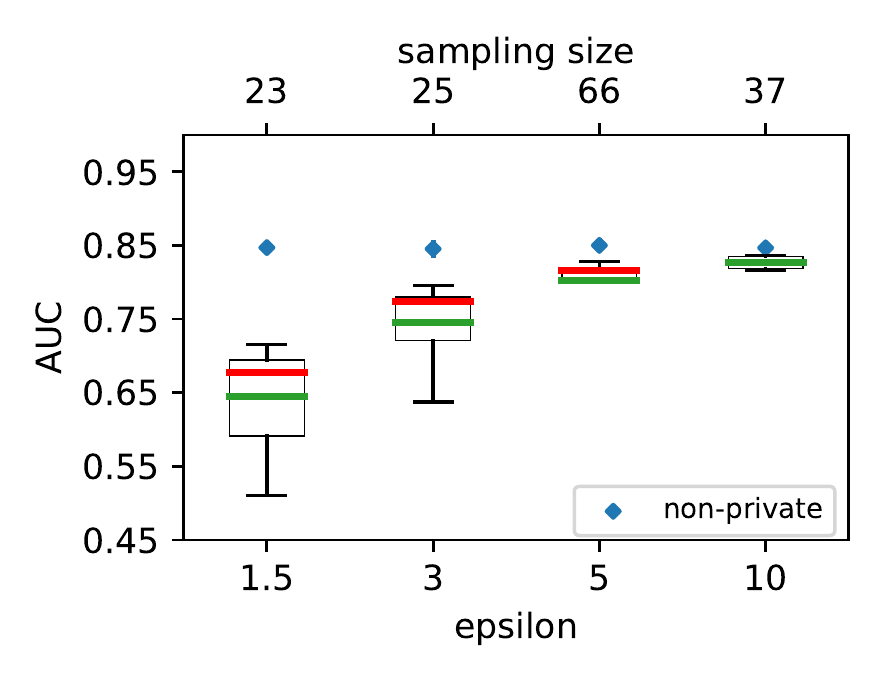}
         \caption{ 
         }
         \label{fig:epsilon_auc}
     \end{subfigure}
     \hfill
     \begin{subfigure}[b]{0.32\textwidth}
         \centering
         \includegraphics[width=\textwidth]{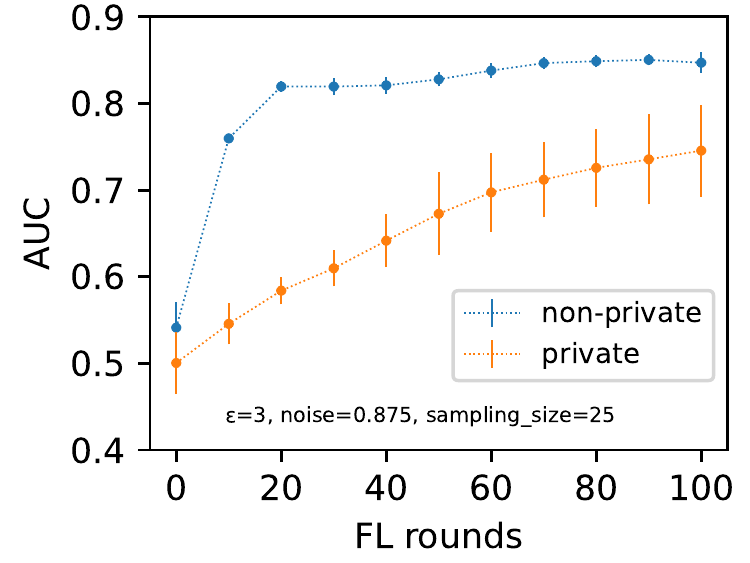}
         \caption{}
         \label{fig:epsilon_3_auc}
     \end{subfigure}
     \hfill
     \begin{subfigure}[b]{0.32\textwidth}
         \centering
         \includegraphics[width=\textwidth]{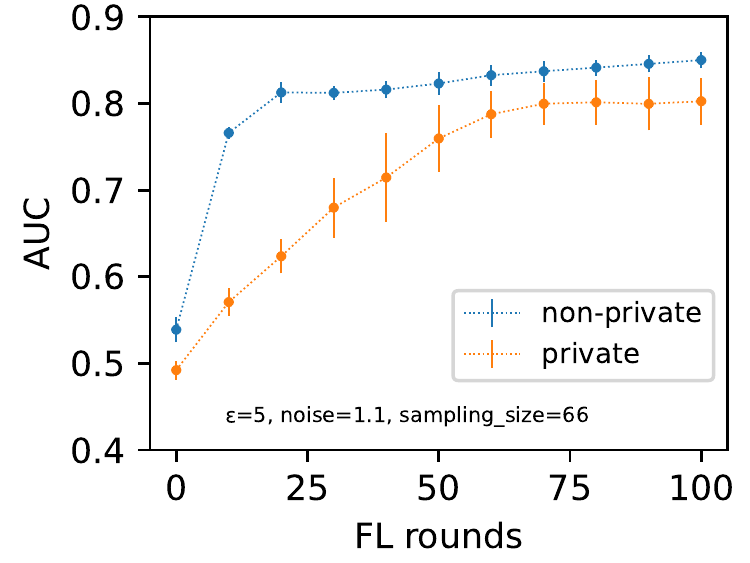}
         \caption{}
         \label{fig:epsilon_5_auc}
     \end{subfigure}
        \caption{Comparing DP and non-DP FL. Evaluation of $(\varepsilon,\delta)$-DP FL for different $\varepsilon$ values and $\delta=10^{-3}$. Experiments with 628 total clients; 100 data points per client; 50\% harmful-class (balanced data). For the non-DP FL, we perform client selection (per FL round) with the same sampling values used for the DP FL.}
        \label{fig:DP_plots}
\end{figure*}

Central Differential Privacy provides privacy guarantees (at the user level) against data (or membership) inference attacks by an external attacker who has access to the trained model. We apply the CDP to our FL training setup (our implementation is based on the TensorFlow privacy library\footnote{\url{https://github.com/tensorflow/privacy}}. TensorFlow modifies the Federated Averaging algorithm to add CDP based on the study of \cite{NEURIPS2021_91cff01a}. The modifications are the following:
(i) each client clips the model's updates before transmitting them to the server.
(ii) the server, during the aggregation of the client's updates, adds noise to the sum of the updates before averaging. 

TensorFlow privacy library provides an implementation that returns the necessary DP parameters (i.e., noise multiplier, sampling size) to achieve a specific $(\varepsilon,\delta)$-DP for the FL training setup. This implementation is based on the Moment Accountant method~\cite{rdpaccountant} which assesses the $(\varepsilon,\delta)$-DP of the model.
Lower $\varepsilon$ values mean that we offer higher privacy to the clients participating in the FL training. The noise multiplier property defines the addition of noise to the sum of the model's updates, and the sampling size refers to randomly selecting a subset of the available clients to participate in each round. The client sampling adds to the privacy guarantee of the training since we do not set a fixed number of clients participating in every round.

We run an experiment to assess the privacy guarantee and utility trade-off. For this experiment, we use the Abusive dataset, split and distribute the data to clients as described in Section~\ref{sec:distributing_dataset}. We run the \textit{FL training setup} for 100 rounds, and each client has a 0.1K balanced dataset. These FL parameters give the maximum available number of clients, i.e, 628 clients. We use Poisson sampling, which gives a different number of clients to participate in each round, with a mean set to \textit{sampling size} value. 

We evaluate the DP-classifier with different $\varepsilon$ values, while setting ${\delta={1e-3}}$. We define $\delta$ using the suggested formula ${\delta=1/|total samples|}$ in~\cite{Abadi_2016, https://doi.org/10.48550/arxiv.1710.06963}. For each $\varepsilon$ value, we get the DP-parameters -- necessary for achieving the given $(\varepsilon, \delta)$-DP -- using the TensorFlow privacy library. So for $\varepsilon$ value of [1.5, 3, 5, 10], we get the following DP-parameters \{sampling size, noise multiplier\}: 1.5=\{23, 1.15\}, 3=\{25, 0.875\}, 5=\{66, 1.1\}, 10=\{37, 0.612\} respectively. We repeated the simulations ten times for $\varepsilon$ set to [1.5, 3], and five times for $\varepsilon$ set to [5, 10]. We present the average AUC achieved in Figure~\ref{fig:epsilon_auc} (the green line shows the mean, and the red the median). 

To investigate the trade-off between utility and privacy, we run a set of experiments with the \textit{FL training setup} using the same parameters mentioned before (i.e., clients dataset, sampling size, number of FL rounds) but without adding DP. In Figure~\ref{fig:epsilon_auc}, we present the average AUC values (over five repetitions) for the non-DP model.
We evaluated the model's performance every ten rounds of the FL training for both the non-DP model and DP model for the $\varepsilon$ values 3 (medium) and 5 (medium-high). We present the average AUC values in Figure~\ref{fig:epsilon_3_auc}, and~\ref{fig:epsilon_5_auc} respectively.

\textit{\textbf{Results Discussion:}}
Figure~\ref{fig:epsilon_auc} shows that adding DP with a strict privacy guarantee (i.e., $\varepsilon=1.5$) causes a 20\% decrease in AUC when compared to the non-DP model performance. Experimenting with lower $\varepsilon$ values, we observed that we do not get a robust model with stable behavior (i.e., four out of ten repetitions gave a 10\% to 30\% AUC). Additionally, we observed that the classifier could tolerate a noise multiplier near the value 1; adding more noise does not allow the classifier to learn during the training. With a medium DP level, $(\varepsilon=3, 5)$, we get an average AUC of 75\%, and 80\%, approaching the non-DP model's performance. Figures~\ref{fig:epsilon_3_auc}, \ref{fig:epsilon_5_auc} show that a DP-model training requires more FL rounds to converge (i.e., 100 rounds) while the non-DP model's performance shows a rapid increase, and reaches an acceptable AUC (i.e., 20-30 rounds). The performance of the non-private model additionally confirms our previous observations that altering the number of FL participants (i.e., sampling size) does not affect the model's performance. Finally, we observe that by training the model for 100 FL rounds, we get 85\% AUC. In other words, the performance is improved by 4\% from the case we present in Figure~\ref{fig:client-size} --- i.e. fifty clients with 0.1K balanced dataset each.

\subsection{Overhead on Client's Device}\label{sec:overhead}

We experiment to measure the extra overhead caused to the client's device when participating in the FL training. Specifically, we assess the overhead during the local training, which happens in one FL round on the client's device.

\begin{figure}[h!]
    \centering
    \includegraphics[width=0.98\columnwidth,height=6cm]{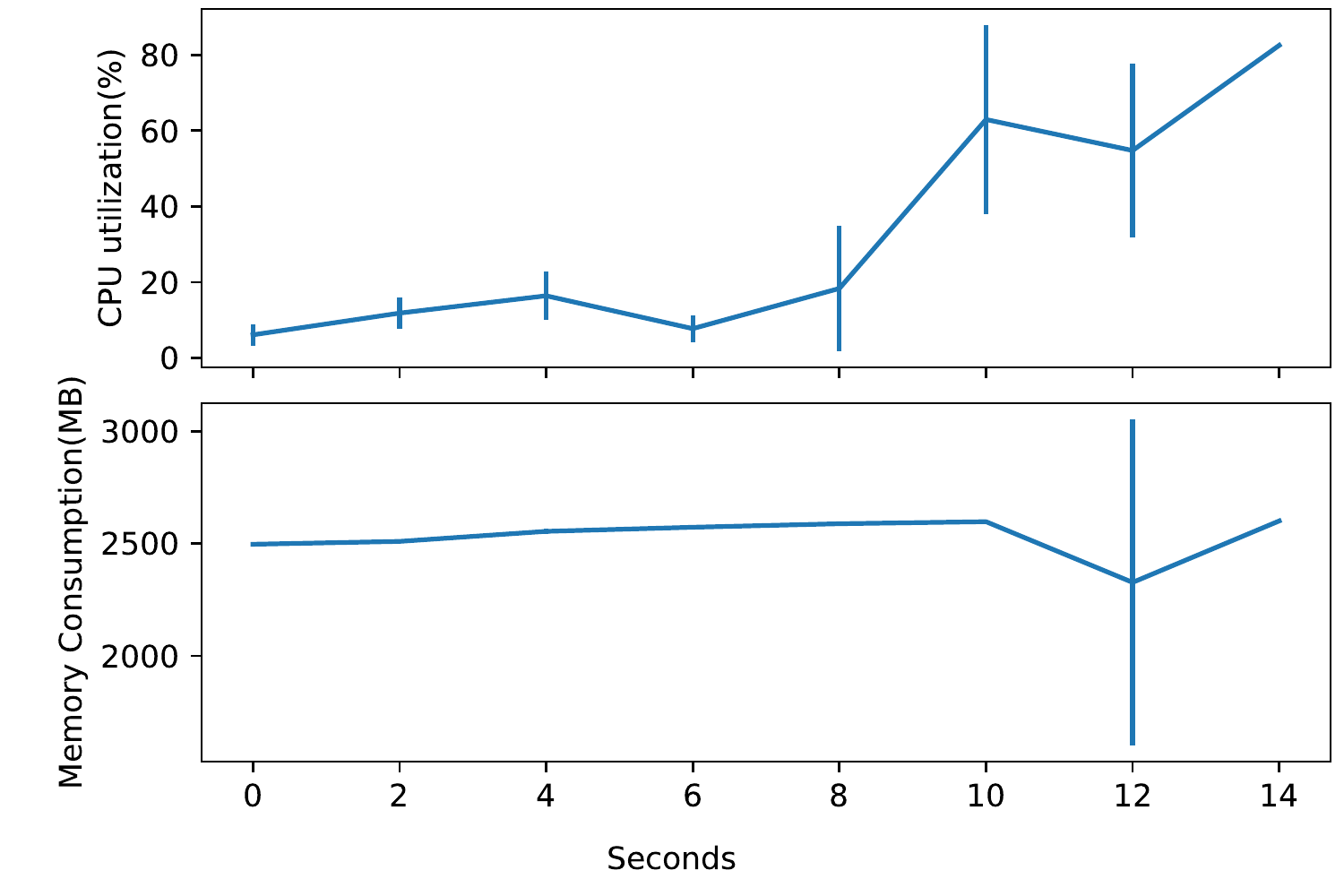}
    \caption{CPU and memory consumption (every 2 seconds) on client's device due to the FL training. The client's training set consists of 0.1K data.}
    \label{fig:memory_cpu}
\end{figure}

We run the \textit{Local training} on a laptop (see laptop properties in Section~\ref{sec:training_setup}), using the whole client's dataset as the training set. Since the results of the experiments with 0.1K data per client showed that we can have a well-performing classifier, we set the client's dataset size to 0.1K.
While training the model locally, we monitor the machine resource utilization (memory consumption and CPU utilization) and collect the logs after every two seconds. We repeated the training ten times. We kept the CPU `idle' during the training by not running other applications. Figure~\ref{fig:memory_cpu} shows the device's CPU utilization percentage and the memory consumption (in MB) during the local training after averaging the results of the experiment.

\textit{\textbf{Results discussion:}}
In Figure~\ref{fig:memory_cpu}, we see that the total duration of the training phase is ${\sim}{14}$ seconds. From seconds 0 to 8, the CPU utilization increases linearly from ${\sim}{10}\%$ to 20\%. Then, there is a rapid increase (from seconds 8 to 10) in which the CPU reaches ${\sim}{70}\%$. At the end of the training phase, there is a decrease to ${\sim}{60}\%$, and CPU utilization reaches a maximum of ${\sim}{80}\%$. The average CPU utilization during the training across all repetitions is ${\sim}{25.5}\%$.

The memory consumption varies between ${\sim}{2300}$ to ${\sim}{2600}$MB during the training, with an average of ${\sim}{2560}$MB. There is a warm-up phase (from 0 to 10) (when the training phase begins) where the memory consumption increases by $\sim$100MB. There is a decrease in memory consumption at 12 seconds (as it also happens with CPU utilization), resulting from one of the repetitions completing the training faster than the rest. Overall, the results show that the local training, with a mean of the CPU utilization around ${\sim}{64}\%$ and at a maximum of ${\sim}{85}\%$, occupies the device for a short time of ${\sim}{14}$ seconds thus, it does not introduce a severe overhead for the client device.

\section{Conclusion}
People of all ages excessively use Online Social Networks and often are exposed to harmful content and various types of misbehavior (i.e., hate speech, cyberbullying, sarcasm, offense, etc.). Online content moderation tools provide countermeasures against such distorted content but at the same time require processing sensitive users' data. The FL paradigm, together with Differential privacy techniques, provides a distributed and private-preserving ML training framework that complies with privacy policies (i.e., GDPR).

In this work, we proposed a privacy-preserving (DP) FL framework for content moderation on Twitter. This DP FL paradigm protects the users' privacy and can be easily adapted to other social media platforms and other types of misbehavior. The experimental results -- over five Twitter datasets -- show that (i) for both the DP and non-DP FL variations, the text classification performance is close to the centralized approach; (ii) it has a high performance even if only a small number of clients (with small local datasets) are available for the FL training; (iii) it does not affect the performance of user’s device -- in terms of CPU and memory consumption -- during the FL training. 

\section{Acknowledgments}
This project has been funded by the European Union’s Horizon 2020 Research and Innovation program under the Cybersecurity CONCORDIA project (Grant Agreement No. 830927) and the Marie Skłodowska–Curie AERAS project (Grant Agreement No. 872735).

\bibliography{paper.bib}

\section*{Ethical Considerations}
This work followed the principles and guidelines on executing ethical information research and using shared data~\cite{ethics}. The suggested methodology complies with the GDPR and ePrivacy regulations. We have not collected data from Twitter. We use existing Twitter datasets -- that have already been published by other academic studies by requesting access from their publishers. For this reason, we will not publicly release any dataset used in this study. We did not use or present any identifiable user information from the datasets (e.g., Twitter user IDs). We applied text preprocessing in order to clean the tweets from any information that could identify specific Twitter accounts (see Section \ref{sec:datasets}). Hence, the train data of the text classifier did not contain Twitter usernames. Finally, we implemented and executed the experiments locally -- on our devices -- without using any cloud computation services, so we did not upload any of the datasets to the cloud.

\end{document}